\pdfoutput=1
\documentclass[11pt,a4paper]{article}
\usepackage[hyperref]{acl2021}
\usepackage{times}
\usepackage{latexsym}

\usepackage{multirow}
\usepackage{multibib}
\usepackage{booktabs}

\usepackage{color}
\usepackage{tikz}
\usepackage{pgfplots}
\usepackage{graphicx}
\usepackage{subfig}
\usepackage{caption}
\usepackage{amsmath}
\usepackage{makecell}

\definecolor{ugreen}{rgb}{0.,0.5,0.}
\usetikzlibrary{backgrounds,fit}

\usepackage{microtype}

\aclfinalcopy 

\title{The NiuTrans End-to-End Speech Translation System \\for IWSLT 2021 Offline Task}

\author{Chen Xu$^1$,
  Xiaoqian Liu$^1$,
  Xiaowen Liu$^1$,
  Laohu Wang$^1$,
  Canan Huang$^1$, \\
  \textbf{Tong Xiao$^{1,2}$},
  \textbf{Jingbo Zhu$^{1,2}$}\\
$^{1}$NLP Lab, School of Computer Science and Engineering\\ 
Northeastern University, Shenyang, China\\
$^{2}$NiuTrans Research, Shenyang, China \\
{\tt
      \{xuchenneu,liuxiaoqianneu,liuxiaowenneu\}@outlook.com,
}\\
{\tt
  \{tigerneu,huangcananneu\}@outlook.com,
}\\
{\tt
      \{xiaotong,zhujingbo\}@mail.neu.edu.cn
}
}
\date{}

\begin{document}
\maketitle
\begin{abstract}

This paper describes the submission of the NiuTrans end-to-end speech translation system for the IWSLT 2021 offline task, which translates from the English audio to German text directly without intermediate transcription.
We use the Transformer-based model architecture and enhance it by Conformer, relative position encoding, and stacked acoustic and textual encoding.
To augment the training data, the English transcriptions are translated to German translations.
Finally, we employ ensemble decoding to integrate the predictions from several models trained with the different datasets.
Combining these techniques, we achieve 33.84 BLEU points on the MuST-C En-De test set, which shows the enormous potential of the end-to-end model.

\end{abstract}

\section{Introduction}

Speech translation (ST) aims to learn models that can predict, given some speech in the source language, the translation into the target language.
End-to-end (E2E) approaches have become popular recently for its ability to free designers from cascading different systems and shorten the pipeline of translation \cite{Duong_naacl2016,Berard_arxiv2016,Weiss_ISCA2017}. 
This paper describes the submission of the NiuTrans E2E ST system for the IWSLT 2021 \cite{iwslt2021} offline task, which translates from the English audio to the German text translation directly without intermediate transcription. 

Our baseline model is based on the DLCL Transformer \cite{Vaswani_nips2017,Wang:ACL2019} model with Connectionist Temporal Classification (CTC) \cite{Graves_ACL2006} loss on the encoders \cite{Bahar_ASRU2019}.
We enhance it with the superior model architecture Conformer \cite{Gulati_ISCA2020}, relative position encoding (RPE) \cite{Shaw:NAACL2019}, and stacked acoustic and textual encoding (SATE) \cite{Xu_CoRR2021}.
To augment the training data, the English transcriptions of the automatic speech recognition (ASR) and speech translation corpora are translated to the German translation.
Finally, we employ the ensemble decoding method to integrate the predictions from multiple models \cite{Wang:WMT18} trained with the different datasets.

This paper is structured as follows. The training data is summarized in Section 2, then we describe the model architecture in Section 3 and data augmentation in Section 4. 
We present the ensemble decoding method in Section 5.
The experimental settings and final results are shown in Section 6.

\section{Training Data}

Our system is built under the constraint condition. 
The training data can be divided into three categories: ASR, MT, and ST corpora\footnote{We only described the training data used in our system.}. 

\noindent \textbf{ASR corpora.} ASR corpora are used to generate synthetic speech translation data. 
We only use the Common Voice \cite{Ardila:LREC2020} and LibriSpeech \cite{Panayotov:ICASSP2015} corpora.
Furthermore, we filter the noisy training data in the Common Voice corpus by force decoding and keep 1 million utterances.

\noindent \textbf{MT corpora}. Machine translation (MT) corpora are used to translate the English transcription. 
We use the allowed English-German translation data from WMT 2020 \cite{Barrault:WMT2020} and OpenSubtitles2018 \cite{Lison:LREC2016}.
We filter the training bilingual data followed \newcite{Li:WMT2019}, which includes length ratio, language detection, and so on.

\noindent \textbf{ST corpora}. The ST corpora we used include MuST-C \cite{Gangi_NAACL2019} English-German\footnote{We use the latest MusST-C v2 dataset released by IWSLT 2021.}, CoVoST \cite{Wang:Corr2020}, Speech-Translation TED corpus\footnote{http://i13pc106.ira.uka.de/˜mmueller/
iwslt-corpus.zip}, and Europarl-ST \cite{Javier_ICASSP2020}. 

The statistics of the final training data are shown in Table \ref{data_stat}.
We augment the quantity of the ST training data by translating the English transcription (the details are unveiled in Section \ref{data_augment}).

\begin{table}[h]
  \centering
  \begin{tabular}{l|l|r|r}
      \toprule
      Task & Corpora & Size & Time \\
      \midrule
      \multirow{3}{*}{ASR} & LibriSpeech & 281241 & 960h \\
      & Common Voice & 1000000 & 1387h \\
      \cmidrule(l){2-4}
      & Total & 1281241 & 2347h \\
      \midrule
      \multirow{6}{*}{MT} & CommonCrawl & 2014304 & - \\
       & Europarl & 1802849 & - \\
       & ParaCrawl & 31528317 & - \\
       & Wiki & 5714363 & - \\
       & OpenSubtitles & 14449099 & - \\
      \cmidrule(l){2-4}
      & Total & 55508932 & - \\
      \midrule
      \multirow{5}{*}{ST} & MuST-C & 249462 & 435h \\
      & CoVoST & 289411 & 329h \\
      & ST TED & 170133 & 254h \\
      & Europarl & 69537 & 155h \\
      \cmidrule(l){2-4}
      & Total & 778543 & 1173h \\
      \bottomrule
  \end{tabular}
  \caption{Data statistics of the ASR, MT, and ST corpora.}
  \label{data_stat}
\end{table}

\section{Model Architecture}

In this section, we describe the baseline model and the architecture improvements. 
Then, the experimental results are shown to demonstrate the effectiveness.

\subsection{Baseline Model}
\label{3.1}

Our system is based on deep Transformer \cite{Vaswani_nips2017} implemented on the fairseq toolkit \cite{Ott:NAACL2019}. 
Furthermore, dynamic linear combination of layers (DLCL) \cite{Wang:ACL2019} method is employed to train the deep model effectively \cite{Li_AAAI2021,Li_EMNLP2020}.

To reduce the computational cost, the input speech features are processed by two convolutional layers, which have a stride of $2$. 
This downsamples the sequence by a factor of 4 \cite{Weiss_ISCA2017}.
For strong systems, we use Connectionist Temporal Classification (CTC) \cite{Graves_ACL2006} as the auxiliary loss on the encoders\cite{Watanabe_IEEE2017,Karita_ISCA2019,Bahar_ASRU2019}.
The weight of CTC objective $\alpha$ is set to 0.3 for all ASR and ST models.
The model architecture is showed in Figure 1\footnote{https://github.com/NiuTrans/MTBook}.

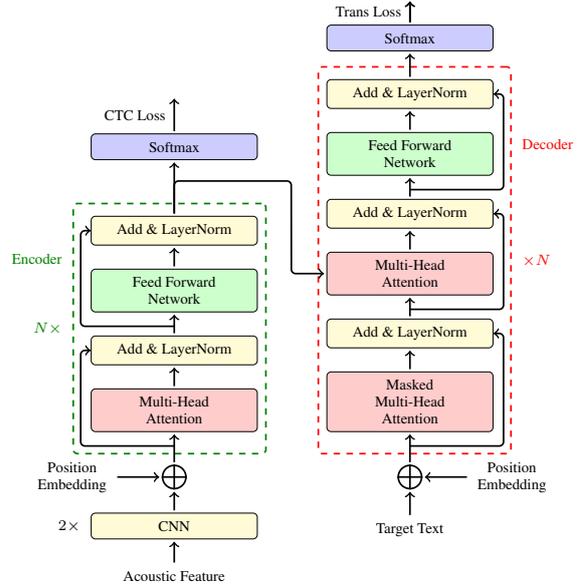
\begin{figure}[t]
  \centering
  \scalebox{.8}{\begin{tikzpicture}
	\tikzstyle{layer}=[draw,rounded corners=2pt,font=\scriptsize,align=center,minimum width=7.1em]
	\tikzstyle{word}=[font=\scriptsize]
\node[layer,fill=red!20] (en_sa) at (0,0){Multi-Head \\ Attention};
\node[anchor=south,layer,fill=yellow!20](en_add1) at ([yshift=1.0em]en_sa.north) {Add \& LayerNorm};
\node[layer,anchor=south,fill=green!20] (en_ffn) at ([yshift=1.0em]en_add1.north){Feed Forward \\ Network};
\node[anchor=south,layer,fill=yellow!20](en_add2) at ([yshift=1.0em]en_ffn.north) {Add \& LayerNorm};
\node[layer,anchor=south,fill=blue!20] (en_sf) at ([yshift=2.4em]en_add2.north){Softmax};
\node[draw,circle,inner sep=0pt, minimum size=1em,anchor=north,thick] (en_add) at ([yshift=-1.4em]en_sa.south){};
\draw[thick] (en_add.90) -- (en_add.-90);
\draw[thick] (en_add.0) -- (en_add.180);
\node[layer,anchor=north,fill=yellow!20] (en_cnn) at ([yshift=-1.0em]en_add.south){CNN};
\node[anchor=east,font=\scriptsize,align=center] (en_pos) at ([xshift=-2em]en_add.west){Position \\ Embedding};
\node[anchor=north,font=\scriptsize,align=center] (en_input) at ([yshift=-1em]en_cnn.south){Acoustic Feature};

\draw[->,thick] (en_input.90) -- ([yshift=-0.1em]en_cnn.-90);
\draw[->,thick] ([yshift=0.1em]en_cnn.90) -- ([yshift=-0.1em]en_add.-90);
\draw[->,thick] ([yshift=0.1em]en_add.90) -- ([yshift=-0.1em]en_sa.-90);
\draw[->,thick] ([yshift=0.1em]en_sa.90) -- ([yshift=-0.1em]en_add1.-90);
\draw[->,thick] ([yshift=0.1em]en_add1.90) -- ([yshift=-0.1em]en_ffn.-90);
\draw[->,thick] ([yshift=0.1em]en_ffn.90) --([yshift=-0.1em]en_add2.-90);
\draw[->,thick] ([yshift=0.1em]en_add2.90) -- ([yshift=-0.1em]en_sf.-90);
\draw[->,thick] ([yshift=0.1em]en_sf.90) -- ([yshift=1.5em]en_sf.90)node[left,pos=0.5]{\scriptsize{CTC Loss}};
\draw[->,rounded corners=2pt,thick] ([yshift=-0.6em]en_sa.south)--([yshift=-0.6em,xshift=-4.0em]en_sa.south)--([xshift=-0.43em]en_add1.west)--(en_add1.west);
\draw[->,rounded corners=2pt,thick] ([yshift=-0.6em]en_ffn.south)--([yshift=-0.6em,xshift=-4.0em]en_ffn.south)--([xshift=-0.43em]en_add2.west)--(en_add2.west);

\node[draw,circle,inner sep=0pt, minimum size=1em,anchor=west,thick] (de_add) at ([xshift=9em]en_add.east){};
\draw[thick] (de_add.90) -- (de_add.-90);
\draw[thick] (de_add.0) -- (de_add.180);
\node[layer,anchor=south,fill=red!20] (de_sa) at ([yshift=1.4em]de_add.north){Masked \\Multi-Head\\Attention};
\node[anchor=south,layer,fill=yellow!20](de_add1) at ([yshift=1.0em]de_sa.north) {Add \& LayerNorm};
\node[layer,anchor=south,fill=red!20] (de_ca) at ([yshift=1.0em]de_add1.north){Multi-Head \\ Attention};
\node[anchor=south,layer,fill=yellow!20](de_add2) at ([yshift=1.0em]de_ca.north) {Add \& LayerNorm};
\node[layer,anchor=south,fill=green!20] (de_ffn) at ([yshift=1.0em]de_add2.north){Feed Forward \\ Network};
\node[anchor=south,layer,fill=yellow!20](de_add3) at ([yshift=1.0em]de_ffn.north) {Add \& LayerNorm};
\node[layer,anchor=south,fill=blue!20] (sf) at ([yshift=1.2em]de_add3.north){Softmax};
\node[anchor=north,font=\scriptsize,align=center] (de_input) at ([yshift=-1.1em]de_add.south){Target Text};

\node[anchor=west,font=\scriptsize,align=center] (de_pos) at ([xshift=2em]de_add.east){Position \\Embedding};

\draw[->,thick] (de_input.90) -- ([yshift=-0.1em]de_add.-90);
\draw[->,thick] ([yshift=0.1em]de_add.90) -- ([yshift=-0.1em]de_sa.-90);
\draw[->,thick] ([yshift=0.1em]de_sa.90) -- ([yshift=-0.1em]de_add1.-90);
\draw[->,thick] ([yshift=0.1em]de_add1.90) -- ([yshift=-0.1em]de_ca.-90);
\draw[->,thick] ([yshift=0.1em]de_ca.90) -- ([yshift=-0.1em]de_add2.-90);
\draw[->,thick] ([yshift=0.1em]de_add2.90) -- ([yshift=-0.1em]de_ffn.-90);
\draw[->,thick] ([yshift=0.1em]de_ffn.90) -- ([yshift=-0.1em]de_add3.-90);
\draw[->,thick] ([yshift=0.1em]de_add3.90) -- ([yshift=-0.1em]sf.-90);
\draw[->,thick] ([yshift=0.1em]sf.90) -- ([yshift=1.0em]sf.90);
\draw[->,thick] ([xshift=0.1em]en_pos.0) -- ([xshift=-0.1em]en_add.180);
\draw[->,thick] ([xshift=-0.1em]de_pos.180) -- ([xshift=0.1em]de_add.0);
\draw[->,rounded corners=2pt,thick] ([yshift=-0.6em]de_sa.south)--([yshift=-0.6em,xshift=4.0em]de_sa.south)--([xshift=0.43em]de_add1.east)--(de_add1.east);
\draw[->,rounded corners=2pt,thick] ([yshift=-0.6em]de_ca.south)--([yshift=-0.6em,xshift=4.0em]de_ca.south)--([xshift=0.43em]de_add2.east)--(de_add2.east);
\draw[->,rounded corners=2pt,thick] ([yshift=-0.6em]de_ffn.south)--([yshift=-0.6em,xshift=4.0em]de_ffn.south)--([xshift=0.43em]de_add3.east)--(de_add3.east);
\draw[->,thick] ([yshift=0.1em]sf.90) -- ([yshift=1.0em]sf.90)node[left,pos=0.5]{\scriptsize{Trans Loss}};

\draw[->,rounded corners=2pt,thick] ([yshift=0.1em]en_add2.90) -- ([yshift=1.5em]en_add2.90) -- ([xshift=5.0em,yshift=1.5em]en_add2.90) -- ([xshift=-1.5em]de_ca.west) -- ([xshift=-0.1em]de_ca.west);

\begin{pgfonlayer}{background}
\node[draw=ugreen,rounded corners=2pt,inner xsep=6pt,inner ysep=8pt,dashed,thick,xshift=-0.2em,yshift=-0.2em][fit=(en_add1)(en_add2)(en_sa)(en_ffn)](box1){};
\node[draw=red,rounded corners=2pt,inner xsep=6pt,inner ysep=8pt,dashed,thick,xshift=0.2em,yshift=-0.2em][fit=(de_sa)(de_ca)(de_ffn)(de_add3)](box2){};
\end{pgfonlayer}

\node[anchor=east,font=\scriptsize,text=ugreen] at ([xshift=-0.1em]box1.west){$N \times$};
\node[anchor=west,font=\scriptsize,text=red] at ([xshift=0.1em]box2.east){$\times N$};
\node[anchor=east,font=\scriptsize] at ([xshift=-0.1em]en_cnn.west){$2 \times$};
\node[anchor=east,font=\scriptsize,align=center,text=ugreen] at ([xshift=-0.1em,yshift=3em]box1.west){Encoder};
\node[anchor=west,font=\scriptsize,align=center,text=red] at ([xshift=0.1em,yshift=5em]box2.east){Decoder};
\end{tikzpicture}}
  \setlength{\abovecaptionskip}{-0.2em}
  \caption{The baseline model architecture.}
\end{figure}
 
\subsection{Conformer}

Conformer \cite{Gulati_ISCA2020} models both local and global dependencies by combining the Convolutional Neural Network and Transformers. It has shown superiority and achieved promising results in ASR tasks.

We replace the Transformer blocks in the encoder by the conformer blocks, which include two macaron-like feed-forward networks, multi-head self attention modules, and convolution modules. 
Note that we use the RPE proposed in \newcite{Shaw:NAACL2019} rather than Transformer-XL \cite{Dai:ACL2019}. 

\subsection{Relative Position Encoding}
\label{rpe}

Due to the non-sequential modeling of the original self attention modules, the vanilla Transformer employs the position embedding by a deterministic sinusoidal function to indicate the absolute position of each input element \cite{Vaswani_nips2017}.
However, this scheme is far from ideal for acoustic modeling \cite{Pham:ISCA2020}.

The latest work \cite{Pham:ISCA2020,Gulati_ISCA2020} points out that the relative position encoding enables the model to generalize better for the unseen sequence lengths.
It yields a significant improvement on the acoustic modeling tasks.
We re-implement the relative position encoding scheme \cite{Shaw:NAACL2019}. 
The maximum relative position is set to 100 for the encoder and 20 for the decoder.
We use both absolute and relative positional representations simultaneously.

\subsection{Stacked Acoustic and Textual Encoding}

The previous work \cite{Bahar_ASRU2019} employs the CTC loss on the top layer of the encoder, which forces the encoders to learn soft alignments between speech and transcription.
However, the CTC loss demonstrates strong preference for locally attentive models, which is inconsistent with the ST model \cite{Xu_CoRR2021}.

In our systems, we use the stacked acoustic-and-textual encoding (SATE) \cite{Xu_CoRR2021} method to encode the speech features. 
It calculates the CTC loss based on the hidden states of the intermediate layer rather than the top layer.
The layers below CTC also extract the acoustic representation like an ASR encoder, while the upper layers with no CTC encode the global representation for translation.
An adaptor layer is introduced to bridge the acoustic and textual encoding.

\subsection{Experimental Results}

\begin{table}[t]
  \centering
  \begin{tabular}{l|c}
      \toprule
      Model & tst-COMMON \\
      \midrule
      Baseline & 23.98 \\
      + Conformer & 24.43 \\
      + RPE & 24.69 \\
      + SATE & 25.35 \\
      \bottomrule
  \end{tabular}
  \caption{Effects of the architecture improvements. We report SacreBLEU scores [$\%$] on the MuST-C tst-COMMON set.}
  \label{arch}
\end{table}

We use the architecture described in Section \ref{3.1} as the baseline model.
The encoder consists of 12 layers and the decoder consists of 6 layers. 
Each layer comprises 256 hidden units, 4 attention heads, and 2048 feed-forward size.
The encoder of SATE includes an acoustic encoder of 8 layers and a textual encoder of 4 layers.
The model is trained with MuST-C English-German dataset and we test the results on the tst-COMMON set based on the SacreBLEU. 
The other experimental details are shown in Section \ref{exp}.

We report the experimental results after applying each architecture improvement in Table \ref{arch}.
Benefitting the power of the deep Transformer, our baseline model achieves 23.98 BLEU points.
The Conformer and RPE methods strengthen the encoding and achieve an improvement of 0.45 and 0.26 BLEU points.
SATE achieves a remarkable improvement by encoding the acoustic representation and textual representation respectively.
We will explore better architecture designs in the future.

\section{Data Augmentation}
\label{data_augment}

\begin{table}[t]
  \centering
  \begin{tabular}{l|l|r|r}
      \toprule
      Data & Corpora & Size & Time \\
      \midrule
      \multirow{5}{*}{Synthetic} & LibriSpeech & 281241 & 960h \\
      & Common Voice & 1000000 & 1387h \\
      & MuST-C & 249462 & 435h \\
      & ST TED & 170133 & 254h \\ 
      \cmidrule(l){2-4}
      & Total & 1700836 & 3036h \\
      \midrule
      Real & Total & 778543 & 1173h \\
      \midrule
      \multicolumn{2}{c|}{Total} & 2479379 & 4209h \\
      \bottomrule
  \end{tabular}
  \caption{All available ST corpora.}
  \label{all_data}
\end{table}

A large amount of the training data is necessary for a strong neural model.
However, unlike the ASR and MT tasks, annotated speech-to-translation data is scarce, which prevents well-trained ST models.
This is the main reason why cascaded systems are the dominant approach in the industrial scenarios.
In this section, we describe our data augmentation method.

We train a deep DLCL Transformer \cite{Wang:ACL2019} with the 25 encoder layers on all available MT data.
To keep the domain consistency with the original ST data, we finetune the MT model on the MuST-C dataset.
The model achieves the SacreBLEU of 35.89 of the MuST-C tst-COMMON test set.
For the case-insensitive LibriSpeech dataset, we train a similar MT model except for lower-casing the source text without punctuation during training.

Then, we generate the German translation from English transcription in the LibriSpeech and Common Voice ASR datasets.
Furthermore, sequence-level knowledge distillation \cite{Kim_EMNLP2016} is applied to augment the training data. 
We generate the translation of the MuST-C and Speech-Translation TED ST datasets which are more related to the target domain.

Corrupting the acoustic feature is another data augmentation method, including SpecAugment, speed perturbation, and so on.
SpecAugment \cite{Park_ISCA2019} is a simple data augmentation applied on the input acoustic features.
The time masking and the frequency masking are applied in our systems.
Speed perturbation transforms the audio by a speed rate, which changes the duration of the audio signal.
Limited by the size of GPU resources, we do not use this method.
Compared with the perturbed data, we think the synthetic samples improve the robustness more effectively.
All available ST corpora are shown in Table \ref{all_data}.

\section{Ensemble Decoding}

Ensemble decoding is an effective method to improve performance by integrating the predictions from multiple models.
It has been proved in the WMT competitions \cite{Wang:WMT18,Li:WMT2019}.
In our systems, we train multiple ST models with different training data for diverse ensemble decoding.
The models are chosen based on the performance of the development set.
This leads to a significant improvement over a single model.

\section{Experiments}
\label{exp}

\subsection{Preprocessing}

We remove the utterances with more than 3000 frames or less than 5 frames.
The 80-channel log-mel filterbank features are extracted from the audio file by torchaudio\footnote{https://github.com/pytorch/audio} library.
We use the lower-cased transcriptions without punctuations for CTC loss computation.
We learn SentencePiece\footnote{https://github.com/google/sentencepiece} subword segmentation with a size of 10,000 based on a shared source and target vocabulary for all datasets.

\subsection{Model Settings}

All experiments are implemented based on the fairseq toolkit\footnote{https://github.com/pytorch/fairseq}.
We use Adam optimizer and adopt the default learning schedule in fairseq.
We apply dropout with a rate of 0.1 and label smoothing $\epsilon_{ls} = 0.1$ for regularization.
We also set the activate function dropout to 0.1 and attention dropout to 0.1, which improves the regularization and overcomes the overfitting.

We use the best model architecture that combines all the improvements described in Section 3. 
The encoder includes an acoustic encoder of 12 conformer layers and a textual encoder of 6 transformer layers.
The decoder consists of 6 Transformer layers.
Each layer comprises 512 hidden units, 8 attention heads, and 2048 feed-forward size.
Pre-norm is applied for training a deep model.
The weight of CTC objective $\alpha$ for multitask learning is set to 0.3 for all models.
All the models are trained for 50 epochs on one machine with 8 NVIDIA 2080Ti GPUs.

During inference, we average the model parameters on the final 10 checkpoints.
We use beam search with a beam size of 5 for all models.
The coefficient of length normalization is tuned on the development set. 
We report the case-sensitive SacreBLEU \cite{Post_wmt18} on the MuST-C tst-COMMON set, IWSLT tst2019 and tst2020 test set.

The organizers provide the segmentation of the test sets and allow the participants to use the own segmentation.
We simply use the segmentation provided by the WerRTCVAD\footnote{https://github.com/wiseman/py-webrtcvad} toolkit.

\subsection{Experimental Results}

Firstly, We train the model on all training corpora, including real and synthetic speech-to-translation paired data.
As shown in Table \ref{all}, we achieve a high BLEU on the tst-COMMON test set, but a low performance on the tst2019 test set compared with the previous work \cite{Gaido_IWSLT2020}.
A possible reason is that the data distribution between IWSLT test sets and the synthetic data is different.

\begin{table}[h]
  \centering
  \begin{tabular}{c|c}
      \toprule
      tst-COMMON & tst2019 \\
      \midrule
      32.65 & 14.16 \\
      \bottomrule
  \end{tabular}
  \caption{Performance of the model trained on all training corpora.}
  \label{all}
\end{table}

To verify this assumption, we pick some subsets from the available datasets for training, including MuST-C and ST TED from the real corpora and MuST-C and LibriSpeech from the synthetic corpora.
We present the results in Table \ref{subset}.
Although the performance on the tst-COMMON test set drops by 0.8 BLEU points, the model achieves a reasonable performance on the tst2019 test set.
Furthermore, we finetune the model on the MuST-C dataset with a small learning rate.
This yields a slight improvement.

\begin{table}[h]
  \centering
  \begin{tabular}{l|c|c}
      \toprule
      Model & tst-COMMON & tst2019 \\
      \midrule
      Base & 31.85 & 20.64 \\
      + finetune & 32.31 & 20.73 \\
      \bottomrule
  \end{tabular}
  \caption{Performance of the model trained with the subsets of all available corpora.}
  \label{subset}
\end{table}

We train multiple models with different training data for diverse ensemble decoding.
We select a part of the synthetic corpora randomly, then mix them with the whole real training data.
Finally, we use the ensemble decoding with 6 models for the final results and achieve a substantial improvement over a single model. 
As shown in Table \ref{final}, we achieve an excellent performance of 33.84 BLEU points on the MuST-C En-De tst-COMMON set.

The best end-to-end system of last year achieves 20.1 BLEU points on the tst2019 test set and 21.49 BLEU points on the tst2020 test set with the given segmentation.
We achieve remarkable improvements of 2.58 and 0.31 BLEU points, which demonstrates the superiority of our systems.

There are two references available for tst2021 test set. The TED reference is the original one from the TED website.
Since new regulations for the official regulation lead to translations that are much shorter, they created a second reference translation, called the IWSLT reference.
The final results are based on both references.
We achieve better performance with the own segmentation on the TED reference, which is consistent with the results on the previous test sets. However, the results with the own segmentation are worse on the IWSLT reference. A possible reason is that we do not optimize the segmentation tool for IWSLT test sets. We will explore better segmentation methods in the future.

\begin{table}[t]
  \centering
  \begin{tabular}{c|c|c}
      \toprule
      Test sets & Given & Own \\
      \midrule
      tst-COMMON & 33.84 & - \\
      tst2019 & 22.68 & 23.76 \\
      tst2020 & 21.8 & 22.8 \\
      tst2021$\dagger$ & 19.0 & 19.6 \\
      tst2021$\ddagger$ & 20.7 & 20.6 \\
      tst2021$\star$ & 30.7 & 30.3 \\
      \bottomrule
  \end{tabular}
  \caption{Final results with ensemble decoding. We report the results with given and own segmentation. There are two references on the tst2021 test set: TED reference ($\dagger$) and IWSLT reference ({$\ddagger$}). The final results are based on both references ($\star$) together.}
  \label{final}
\end{table}

\section{Conclusion}

This paper describes the submission of the NiuTrans E2E ST systems for the IWSLT 2021 offline task, which translates the English audio to German translation directly without intermediate transcription. 
We build our final submissions considering two mainstreams:

\begin{itemize}
  \item Model architecture improvements for the speech translation task.
  \item Data augmentation by translating the English transcription to German translation.
\end{itemize}

We also find that the distribution of the training data has a great impact on the performance and alleviate it by ensemble decoding.
Using the given segmentation, we achieve remarkable improvements over the best end-to-end system of last year.

\section{Acknowledgement}

This work was supported in part by the National Science Foundation of China (Nos. 61876035 and 61732005), the National Key R\&D Program of China (No. 2019QY1801), and the Ministry of Science and Technology of the PRC (Nos. 2019YFF0303002 and 2020AAA0107900). The authors would like to thank anonymous reviewers for their comments.

\bibliographystyle{acl_natbib}
\bibliography{acl2021}

\end{document}